\documentclass{article}

\usepackage[preprint, nonatbib]{neurips_2023}

\usepackage[utf8]{inputenc} 
\usepackage[T1]{fontenc}    
\usepackage{hyperref}       
\usepackage{url}            
\usepackage{booktabs}       
\usepackage{amsfonts}       
\usepackage{nicefrac}       
\usepackage{microtype}      
\usepackage{xcolor}         
\usepackage{graphicx}
\usepackage[inkscapeformat=png]{svg}
\usepackage[flushleft]{threeparttable}
\usepackage{adjustbox}
\usepackage{floatrow}
\newfloatcommand{capbtabbox}{table}[][\FBwidth]
\DeclareMarginSet{hangboth}{\setfloatmargins*{\hskip-4cm}{\hskip-4cm}}
\usepackage{wrapfig}

\usepackage{amsmath}
\DeclareMathOperator*{\argmax}{argmax}
\DeclareMathOperator*{\argmin}{argmin}
\def\ci{\perp\!\!\!\perp}

\usepackage[capitalize]{cleveref}
\crefname{section}{Sec.}{Secs.}
\Crefname{section}{Section}{Sections}
\Crefname{table}{Table}{Tables}
\crefname{table}{Tab.}{Tabs.}

\definecolor{myorange}{RGB}{252, 135, 10}
\definecolor{mygreen}{RGB}{0, 156, 26}

\title{OD-NeRF: Efficient Training of On-the-Fly Dynamic Neural Radiance Fields}

\author{%
    Zhiwen Yan \qquad Chen Li \qquad Gim Hee Lee\\
    Department of Computer Science, National University of Singapore\\
    {\tt\small \{yan.zhiwen, lichen\}@u.nus.edu} \qquad {\tt\small gimhee.lee@nus.edu.sg}
}

\begin{document}

\maketitle

\begin{abstract}
    Dynamic neural radiance fields (dynamic NeRFs) have demonstrated impressive results in novel view synthesis on 3D dynamic scenes.
    However, they often require complete video sequences for training followed by novel view synthesis, which is similar to playing back the recording of a dynamic 3D scene. In contrast, we propose OD-NeRF to efficiently train and render dynamic NeRFs on-the-fly which instead is capable of streaming the dynamic scene. When training on-the-fly, the training frames become available sequentially and the model is trained and rendered frame-by-frame. The key challenge of efficient on-the-fly training is how to utilize the radiance field estimated from the previous frames effectively. To tackle this challenge, we propose: 1) a NeRF model conditioned on the multi-view projected colors to implicitly track correspondence between the current and previous frames, and 2) a transition and update algorithm that leverages the occupancy grid from the last frame to sample efficiently at the current frame.
    Our algorithm can achieve an interactive speed of 6FPS training and rendering on synthetic dynamic scenes on-the-fly, and a significant speed-up compared to the state-of-the-art on real-world dynamic scenes. 
\end{abstract}

\section{Introduction}
Neural radiance fields (NeRF) \cite{mildenhall2021nerf} have recently emerged as a new 3D volumetric representation capable of synthesizing photo-realistic novel views from multi-view images. Many dynamic NeRFs are also proposed to reconstruct scenes with moving or deforming objects, using space-time 4D \cite{li2022dynerf, park2023temporal} or explicit deformation \cite{pumarola2020dnerf, park2021hypernerf} representations. However, most of the dynamic NeRFs focus on reconstructing the scene given the full training videos of a dynamic event. Although capable of playing back the dynamic scene after the event, they cannot stream the event while it is happening. In this paper, we introduce a new ``on-the-fly training`` approach to train a dynamic NeRF concurrently with the acquisition of the training frames, as the dynamic event unfolds. For the purpose of streaming the dynamic scene, the model only needs to render the current radiance field anytime during the training.
With the recent success in NeRF acceleration to train \cite{mueller2022instant,Chen2022tensorrf} a small scene in seconds, it becomes a promising possibility to further speed up the dynamic scene on-the-fly training to an interactive speed.
This enables the streaming of dynamic scenes with a wide variety of applications in social media, VR/AR, and gaming industries. 

Training a dynamic NeRF on-the-fly is considerably different from training it after the dynamic event. For training after the event, the NeRF is trained to fit the 4D space-time radiance field with the available full training videos. In contrast, for training on-the-fly, NeRF is trained to represent the current radiance field given the previous reconstruction and the available video frames up to the current time step. As the dynamic event continues, new frames will become available and the NeRF will be updated. Training with each frame from scratch is very time-consuming, so the key to efficient on-the-fly dynamic NeRF training is to effectively utilize the radiance fields estimated from the previous frames for the faster convergence of the current radiance field. By leveraging the correspondence and transition across frames, we propose: 1) a NeRF representation conditioned on multi-view projected colors for faster convergence, and 2) a transition and update to the occupancy grid used for efficient sampling.

Our proposed NeRF representation conditioned on multi-view projected colors is designed for fast convergence when training the radiance fields of consecutive frames on-the-fly. Most of the existing dynamic NeRFs explicitly represent the motion using a temporal input dimension \cite{li2022dynerf,park2023temporal} in NeRF or a temporal deformation field \cite{pumarola2020dnerf, park2021hypernerf}. 
Due to the lack of direct supervision or known dynamics, these models with temporal input often extrapolate poorly to the unseen time and require many iterations of optimization for each frame. Instead of being conditioned on time, we propose a model that is conditioned on the projected colors from the training views. This is based on the observation that the projected color of a corresponding 3D point often stays unchanged in consecutive frames. With guidance from the invariant projected colors, the NeRF model is implicitly aware of the point correspondence across consecutive frames. 
Consequently, the NeRF model can effectively utilize the radiance fields from the previous frames to render for the current frame with the implicit correspondence. 
The experiments suggest that our proposed model has excellent temporal extrapolation capability and requires minimum number of optimization iterations for each new frame. As a result, our on-the-fly training speed of the model is significantly improved.

Additionally, we introduce a method of transiting and updating the occupancy grid used for efficient point sampling. 
Since most of the 3D scenes are empty spaces, the occupancy grid has been used in static NeRFs \cite{mueller2022instant,
Chen2022tensorrf} to reduce the number of sampled points for acceleration. To adapt the occupancy grid to on-the-fly dynamic training, we consider the occupancy grid as a probability of the 3D voxels occupied by any object. To probabilistically model the motions in the scene, a transition function is applied to occupancy probability at the start of each frame optimization and later updated with new observations. The updated occupancy grid can then be used to sample points only in the occupied areas anytime during the on-the-fly training. 

We evaluate our method on synthetic D-NeRF dataset \cite{pumarola2020dnerf}, real-world MeetRoom \cite{li2022streamrf}, and DyNeRF \cite{li2022dynerf} dataset. When trained and rendered on-the-fly, our method achieves significant acceleration compared to the state-of-the-art algorithms while maintaining a comparable rendering quality. Particularly, our method can train and render 6 frames per second (FPS) on the synthetic D-NeRF dataset. 
We summarize \textbf{our contributions} as follows: 1) we introduce and formally formulate the new setting of training dynamic NeRF on-the-fly. 2) We propose a projected color-guided on-the-fly dynamic NeRF and a transiting occupancy grid for efficient on-the-fly training. 3) We achieve $10\times$ on-the-fly training and rendering acceleration in synthetic scenes and $3\times$ acceleration in real-world scenes compared to the state-of-the-art.

\section{Related Works}
\subsection{Accelerated Dynamic Neural Radiance Fields}
Following the success in representing static 3D scenes with Neural Radiance Fields(NeRFs) \cite{mildenhall2021nerf, wang2021neus, yu2021plenoxels}, many works \cite{pumarola2020dnerf, li2022dynerf, park2021hypernerf, yan2023nerfds} have been exploring representing dynamic scenes with NeRFs as well. Similar to the early static NeRFs, they usually take at least hours to train for a single scene. To improve the training efficiency of dynamic NeRFs, some works try to migrate the acceleration methods used for static NeRFs to dynamic NeRFs. TiNeuVox \cite{TiNeuVox} follows the voxel and Multi-Layer Perceptron(MLP) architecture as in \cite{mueller2022instant} to represent static canonical space, and uses another MLP to capture temporal deformation. StreamRF \cite{li2022streamrf} represents the scene at the first frame using the Plenoxel \cite{yu2021plenoxels}, and learns the subsequent changes to this voxel grid. \cite{park2023temporal, kplanes_2023} directly expand the 3D representation used in static NeRF to 4D with a time dimension to represent dynamic scenes. However, most of the existing works only focus on joint training of the dynamic scenes, instead of on-the-fly training introduced in our work. Their per frame training speed is also not fast enough for interactive on-the-fly applications.

\subsection{Image Based Rendering}
Instead of representing a 3D scene as a NeRF conditioned on spatial coordinates and viewing direction, some works rely on additional projected colors/features on the training views to improve generalization or robustness. IBRNet \cite{wang2021ibrnet}, MVSNeRF \cite{mvsnerf} and many following works \cite{huang2023localgeneralizable, ren2022volrecon, long2022sparseneus} construct a cost volume of a dynamic scene based on the image features of the nearby views to learn a blending weights or the density and color output. These models are designed to be generalizable to unseen scenes as they do not rely on the spatial coordinates input. LLFF \cite{suhail2022light, suhail2022gpnr} aggregates features along the multi-view epipolar line to render view dependent effect. Recently, DynIBaR \cite{li2022dynibar} applies this technique to dynamic NeRF by aggregating the projected image features across frames, after warpping a point using a motion trajectory learned from past and future frames. However, this method requires time-consuming optimization of the per-frame motion trajectory and cannot be applied to on-the-fly training as the future frames are not known. 

\section{Problem Formulation}
In this section, we first briefly describe the NeRF preliminaries necessary to understand our formulation. We then formally define our new on-the-fly dynamic NeRF training and followed by simplifying the problem formulation in the form of a Hidden Markov model (HMM). 

\paragraph{NeRF Preliminaries.}
Static neural radiance fields \cite{mildenhall2021nerf} represent a 3D scene implicitly with a continuous volumetric function $F:(\mathbf{x}, \mathbf{d})\mapsto(\sigma, \mathbf{c})$ that maps the 
spatial position $\mathbf{x} \in \mathbb{R}^3$ and 
viewing direction $\mathbf{d} \in \mathbb{R}^3$ to the volume density $\sigma \in \mathbb{R}$ and RGB color $\mathbf{c} \in \mathbb{R}^3$. To synthesize the pixel color $\hat{C}(\mathbf{r})$ on any 2D images, volume rendering is used to aggregate the color of $N$ points with interval $\delta$ along the ray $\mathbf{r}$ shooting from the pixel:
\begin{equation}
    \hat{C}(\mathbf{r})=\sum_{i=0}^{N-1}T_{i}(1-\exp(-\sigma_i\delta_i))\mathbf{c}_i, \quad T_i=\exp(-\sum_{j=0}^{i-1}\sigma_j\delta_j).
\end{equation}
Dynamic NeRFs usually include an additional time dimension $t$ in the input, through either a time-varying deformation field $D:(\mathbf{x}, t)\mapsto(\mathbf{x'})$ that maps the spatial coordinates $\mathbf{x}$ to its canonical space correspondence $\mathbf{x'}$ \cite{pumarola2020dnerf, park2021hypernerf}, or directly expanding the 3D NeRF model to a 4D variant $F':(\mathbf{x}, \mathbf{d}, t)\mapsto(\sigma, \mathbf{c})$ \cite{li2022dynerf, park2023temporal}. During the training stage, all $K$ frames of the training images $C_{0:K}(\mathbf{r})$ from all time $t_{0:K}$ are used to jointly minimize the rendering loss:
\begin{equation}
    \mathcal{L}=\sum_{t=0}^{K}\sum_{\mathbf{r}\in\mathcal{R}}||\hat{C}_t(\mathbf{r})-C_t(\mathbf{r})||_2^2.
\end{equation}

\begin{figure}[t]
    \centering
    \includegraphics[width=\linewidth]{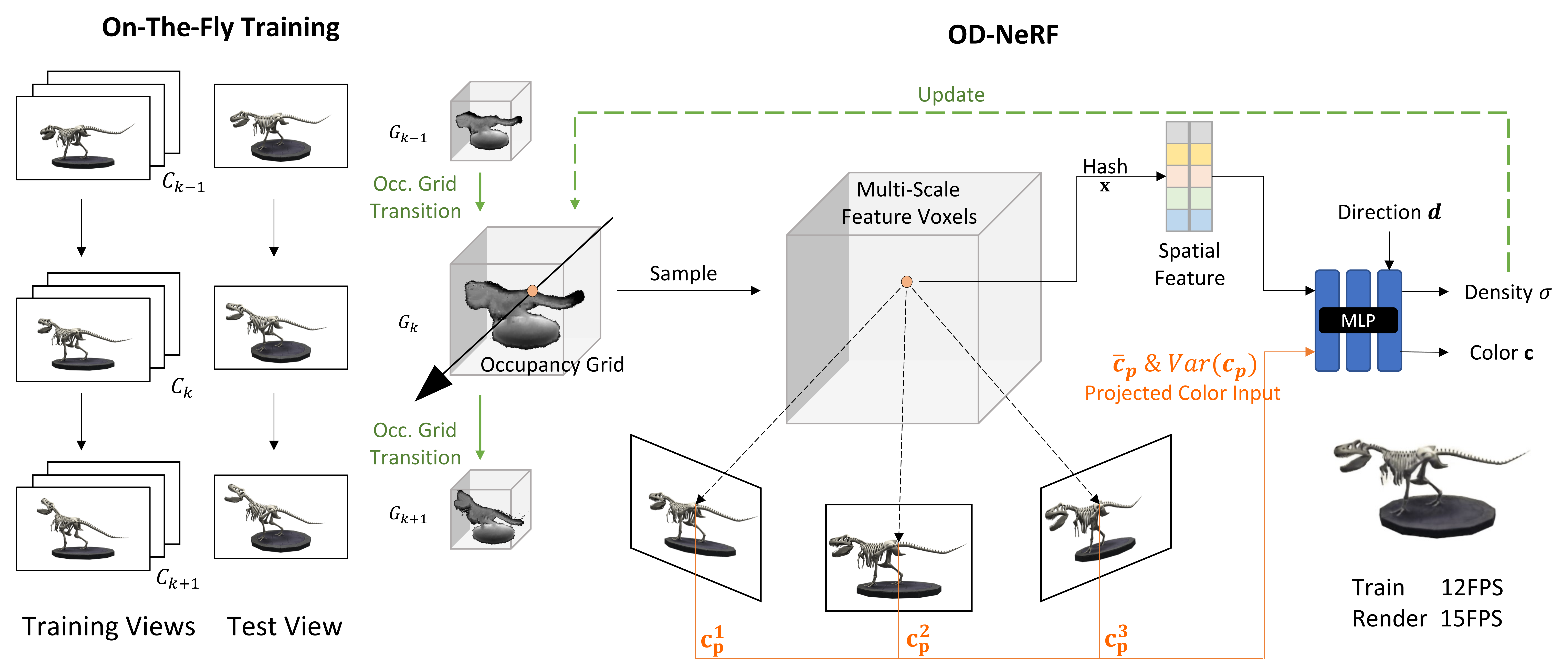}
    \caption{We introduce the on-the-fly training(left) of dynamic NeRFs and the OD-NeRF model(right). In on-the-fly training, the dynamic NeRF is trained based on the current and previous training frames to synthesize novel views for the current time step. Our OD-NeRF leverages the projected colors(\textcolor{myorange}{orange arrow}) to track implicit correspondence for fast on-the-fly convergence, and transition and update to the occupancy grid(\textcolor{mygreen}{green arrows}) for efficient sampling.}
    \label{fig:architecture}
\end{figure}

\paragraph{On-the-fly Dynamic NeRF.} We introduce a new on-the-fly training of the dynamic NeRF 
suitable for streaming dynamic scenes. 
Instead of training after the dynamic event, we train the NeRF concurrently while the dynamic event 
unfolds. 
Furthermore, the NeRF trained at 
time $t_k$ renders 
novel views only at time $t_k$. 
Given the images up to the current time step $C_{0:k}$ and the radiance field estimated up to the last time step $F_{0:k-1}$, the goal of on-the-fly dynamic NeRF training is to find the radiance field function $F_k$ at time $t_k$ that minimizes the rendering loss at the current time step:
\begin{equation}
\begin{split}
    F_k(\mathbf{x}, \mathbf{d})&=\argmax_{F_k} P(F_k \mid C_{0:k}, F_{0:k-1}) \\
                               &=\argmin_{F_k}\sum_{\mathbf{r}\in\mathcal{R}}||\hat{C}_k(\mathbf{r})-C_k(\mathbf{r})||_2^2.
\end{split}
\end{equation}

However, estimating the radiance field $F_k$ at the current time step conditioned on all the previous images $C_{0:k}$ and radiance field $F_{0:k-1}$ is not scalable as time $t_k$ increases. 
To mitigate this growth in complexity, we apply the first order Markov assumption to simplify the probability model by assuming conditional independence $F_k \ci \{C_{0:k-1}, F_{0:k-2} \} \mid \{C_k, F_{k-1}\}$. This simplifies the estimation of the current radiance field as:
\begin{equation}
    F_k=\argmax_{F_k}P(F_k \mid C_k, F_{k-1}).
\end{equation}
Taking the radiance fields $F_{0:K}$ as hidden states and images $C_{0:K}$ as observations, we can formulate the on-the-fly training as the process of estimating the hidden states in a Hidden Markov model (HMM).
The emission function $P(C_k \mid F_k)$ is the process of volumetric rendering that renders 2D images from 3D radiance fields. The transition function $P(F_k \mid F_{k-1})$ is the radiance field deformation or motion between two consecutive time steps. 

Based on the formulation above, the key to efficient on-the-fly training is to maintain low complexity of the update $P(F_k \mid C_k, F_{k-1})$
to the radiance field at each time step.
To this end, we propose a projected color-guided NeRF that is implicitly aware of point correspondence that requires minimum optimization when transiting from $F_{k-1}$ to $F_k$ for fast training. Furthermore, we also introduce a simple transition function $P(G_k|G_{k-1})$ to the occupancy grid $G$ used for efficient sampling. 

\paragraph{Remarks.}
\label{problem_formulation}
Note that we limit the scope of this work to dynamic scenes captured by multi-view forward-facing cameras based on realistic considerations. Although the reconstruction of dynamic scenes from a monocular camera is less demanding on the hardware, it requires the photographer to keep on moving the camera \cite{park2021nerfies}. This is 
cumbersome in prolonged streaming scenarios. 360-degree inward-facing cameras can be used to reconstruct from all angles. Nonetheless, this often requires dozens of cameras and a much bigger space~\cite{peng2021neural}. It is difficult for most streamers to acquire such professional setups. 
Consequently, we focus on scenes captured by static multi-view forward-facing cameras that are most aligned with the setups used in the current streaming industry.

\section{Method}

As mentioned in the previous section, the 
goal of on-the-fly training is to estimate the current radiance field $F_k$ based on the last radiance field $F_{k-1}$ and the current training images $C_{k}$. In practice, this can be achieved by optimizing the radiance field model from the last step with the current training images. 
To achieve highly efficient on-the-fly training, we can either reduce the number of optimization iterations needed for each time step or reduce the time spent on each iteration without scarifying the performance. Based on our HMM-based on-the-fly training paradigm, we propose: 1) a dynamic NeRF guided by projected colors, and 2) an occupancy grid transition and update strategy, to achieve these two goals respectively (\cref{fig:architecture}).

\subsection{Dynamic NeRF Guided by Projected Colors} 

According to the HMM formulation of on-the-fly training, the optimization at each time step is effectively learning a probability update to the last step model given the current observations: $F_k=P(F_k \mid F_{k-1}, C_k)F_{k-1}$. In the context of dynamic NeRFs, this can usually be achieved by estimating a deformation field or point correspondence function. 

\begin{figure}
    \centering
    \includegraphics[width=\linewidth]{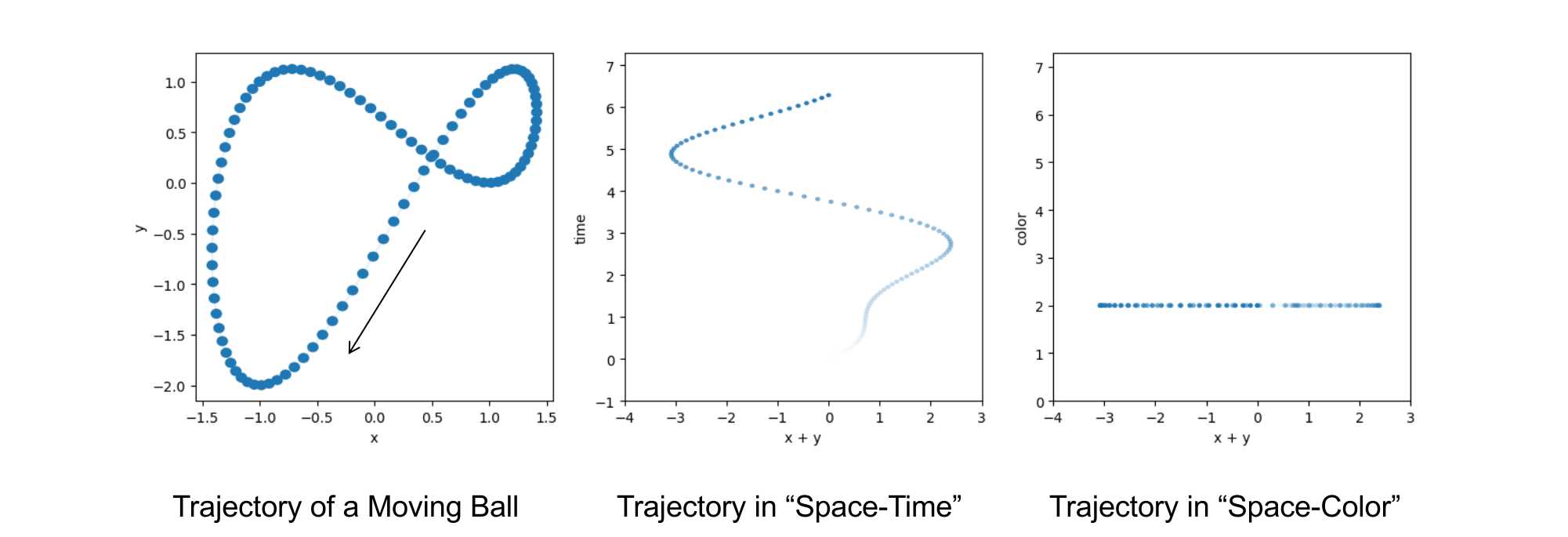}
    \caption{\small We illustrate the trajectory of a blue ball moving on a $(x, y)$ over time(left). When mapping it to the ``space-time'' coordinate system, the point correspondence can be hard to track due to the irregular trajectory(middle, 2D space represented with $x+y$ for simpler visualization). When mapping it to the ``space-color'' coordinate system, the correspondence can be easily tracked due to the invariant color(right). }
    \label{fig:correspondence}
\end{figure}

Existing deformation-based dynamic NeRFs try to learn a deformation field $D:(\mathbf{x}, t_k)\mapsto(x')$ that captures the point correspondence through joint training. However, our experiments indicate that this process itself takes too long for efficient on-the-fly training. We postulate that this is because the temporal deformation field does not have a clear pattern in most cases. We illustrate this problem with a toy example as shown in~\cref{fig:correspondence}. In this example, we track the center of the 2D moving ball of a constant color. When tracking the trajectory of the ball in ``space-time'' input space, it is difficult to accurately predict 
the location of the ball at the next time step. This is similar to the temporal deformation field used by many existing dynamic NeRFs. The deformation parameters need to be optimized at each time step based on the indirect RGB supervision. In the case of 3D neural radiance field reconstruction, it is very time-consuming to optimize for all points in the space. 

To efficiently utilize the point correspondence across frames, we change the time input $t$ in the space-time dynamic NeRF model $F(\mathbf{x}, \mathbf{d}, t)$ to the multi-view projected color mean $\bar{\mathbf{c}}_p(\mathbf{x}, k)$ and variance $\operatorname{Var}(\mathbf{c}_p(\mathbf{x}, k))$ at frame $k$ to become: 
\begin{equation}
    F_k: (\mathbf{x}, \mathbf{d}, \bar{\mathbf{c}}_p(\mathbf{x}, k), \operatorname{Var}(\mathbf{c}_p(\mathbf{x}, k)))\mapsto (\sigma, \mathbf{c}).
\end{equation}
The projected colors $\mathbf{c}_p(\mathbf{x}, k)$ are defined as the pixel colors of the training images $C_{k,\text{cam}}$ of all cameras $\mathcal{M}$ through
3D to 2D projection with camera projection matrix $\mathbf{P}_\text{cam}$:
\begin{equation}
    \mathbf{c}_p(\mathbf{x}, k) = \{C_{k, \text{cam}}(\mathbf{P}_\text{cam}\cdot\mathbf{x}) \mid \text{cam} \in \mathcal{M}\}.
\end{equation}
As the model is conditioned on the spatial coordinates and projected colors, it circumvents the difficulty of estimating the point correspondence with irregularity in the spatial and temporal dimensions. Instead, it is much easier to estimate the correspondence in the spatial and color input space due to the invariance of the point color under motion. As illustrated in~\cref{fig:correspondence}, the trajectory of the point in the spatial and color input space is much more regular than that in the spatial and temporal input space. Although the projected colors of 3D points are not perfectly invariant due to factors such as reflection, lighting, and occlusion, we observe that the projected colors stay largely similar across consecutive frames. 

\begin{figure}
    \centering
    \includegraphics[width=0.8\linewidth]{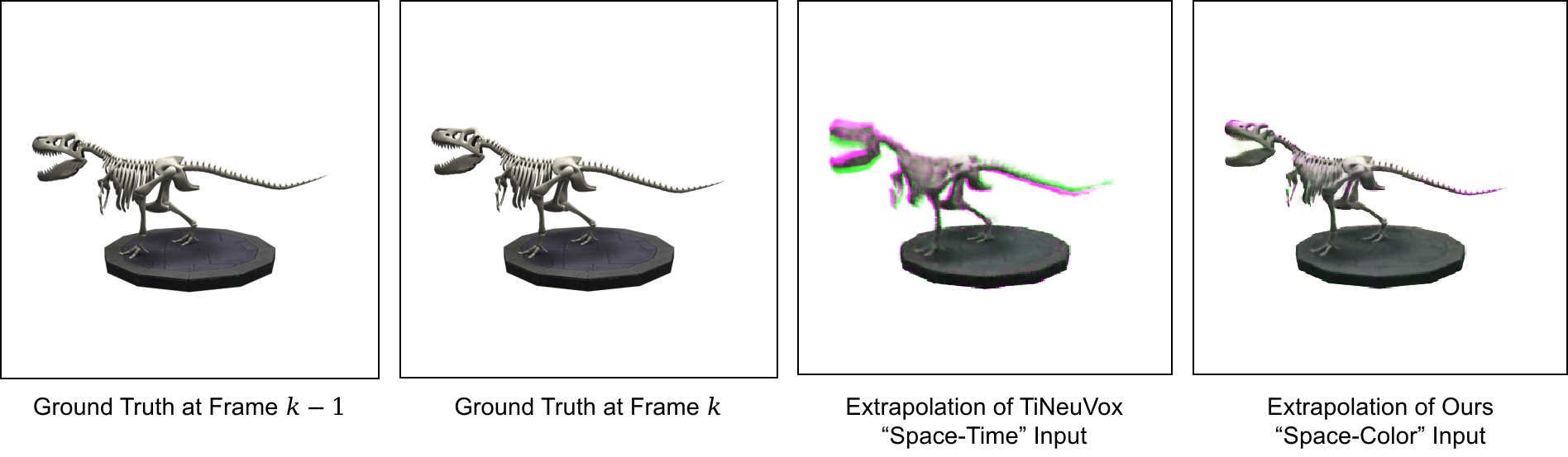}
    \caption{\small We compare the extrapolation ability of dynamic NeRF with ``space-time'' input and with ``space-color'' input. The two images on the right illustrate the novel view synthesized for time $t_k$ by the respective model trained with frames up to $t_{k-1}$. The green channel represents the rendered image, and the purple challenge represents the ground truth for frame $k$. The model with ``space-time'' input extrapolates poorly and renders the image lagging behind the dynamic ground truth, while our model with ``space-color'' input extrapolates well without any training on frame $k$. }
    \label{fig:extrapolation}
\end{figure}

To verify that our method is better at estimating point correspondence on-the-fly, we demonstrate a simple extrapolation experiment in~\cref{fig:extrapolation}. Specifically, we train the different dynamic NeRFs on-the-fly up to the $(k-1)$th frame and then extrapolate to render 
the $k$th frame without any training on the $k$th frame. 
As shown in~\cref{fig:extrapolation}, the model operating in the spatial and temporal input dimensions extrapolates poorly as it is not aware of the point correspondence with untrained time. Our model operating in the spatial and projected color input dimensions can extrapolate well based on the point correspondence hinted by the projected colors. With the implicit point correspondence and the good extrapolation ability, our model requires a 
low number of iterations for each new frame and thus is very fast when training on-the-fly.

\subsection{Occupancy Grid Transition and Update}
Occupancy grids are often used in static NeRFs to reduce the number of points sampled by caching whether a voxel is occupied. 
Formally, the occupancy grid is a 3D voxel grid $G = \{\max(\sigma(\mathbf{x})) \mid \forall \mathbf{x}\in \mathbb{V}_\text{cur}\}^3$, where $\max(\sigma(\mathbf{x}))$ represents the maximum volume density of all points in the respective voxel $\mathbb{V}_\text{cur}$. When sampling points on a ray, only points within the voxel above a certain volume density threshold are kept. It is obvious that this cannot be directly applied to dynamic scenes since the volume density of the 3D space changes over time. 
One way of applying this approach in dynamic scene reconstruction is to maintain a space-time 4D occupancy grid \cite{park2023temporal}. Unfortunately, this does not improve the sampling efficiency when training on-the-fly as the 3D occupancy at $t_{k-1}$ does not affect the occupancy grid at $t_k$. When training the new frame $k$, the occupancy grid at the current time $t_k$ is the same as being initialized from scratch.

\begin{wrapfigure}{r}{0.4\textwidth}
      \ffigbox[\FBwidth]
    {
    \includegraphics[width=\textwidth]{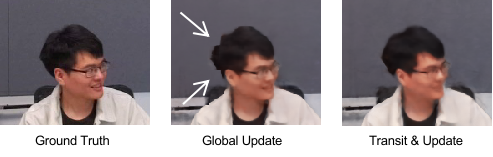}
    }
    {\caption{\small Updating the occupancy grid by sampling random points in each voxel can miss the moving occupied space and causes the entire voxel skipped during rendering. Our transit and update method does not have this issue as it updates the grid at the occupied point. }
    \label{fig:occ_update}
    }
    \vspace{-4mm}
\end{wrapfigure}

To tackle this problem, we follow our Hidden Markov Model formulation to apply a simple transition function and belief update to the occupancy grid. We consider the occupancy grid at any time $t_k$ as a 3D probability function $G_k = \{P(\max(\sigma_k(\mathbf{x})) > 0) \mid \forall \mathbf{x} \in \mathbb{V}_\text{cur}\}^3$, representing the chance of any point present in the voxel with positive volume density at the current time step. Since the occupancy grid $G_k$ 
is constantly updated throughout the per-frame training, we use $G_k^j$ to denote the occupancy grid after $j$ iterations of optimization where $j\in[0, J]$. At the start of each new frame optimization, we apply a transition function to this occupancy grid for the possible motions of the objects in the 3D space. 
This transition function takes the form of a simple 3D convolution kernel $S$ because the occupancy grid is a 3D tensor, such that:
\begin{equation}
    G_k^0=P(G_k \mid G_{k-1})\cdot G_{k-1}^J=S\ast G_{k-1}^J.
\end{equation}
Since we formulate the problem with the Markov assumption for higher efficiency, we have little information about the actual motion based on the previous frames. Thus, we assign the kernel with a simple 3D Gaussian function to represent the probability of the motion. 

After the transition function is applied to the occupancy grid at the start of the optimization for each frame, it needs to be updated with the new observations. Similar to the existing occupancy grid methods, we apply a simple Bayesian update to the occupancy grid probabilities using the volume density output $\sigma(\mathbf{x})$ of the NeRF model:
\begin{equation}
    G_k^j=P(\sigma(\mathbf{x}) \mid G_k^{j-1})\cdot G_k^{j-1}, \quad
    P(\sigma(\mathbf{x}) \mid G)=
    \begin{cases}
        1,                   & \text{if}~~\sigma(\mathbf{x}) < G(\mathbf{x}) \\
        \sigma(\mathbf{x}),  & \text{otherwise}
    \end{cases}.
\end{equation}
Note that the global update method used in Instant-NGP \cite{mueller2022instant} can also be used to update the occupancy grid. However, the global sampling is not suitable for a constantly changing occupancy grid. The global update method randomly samples points in each occupancy voxel to update the occupancy grid. It ignores the previous occupancy values stored and the transition across frames entirely. Since the model takes much fewer iterations to optimize for each new frame when trained on-the-fly, the random sampling in each voxel can miss the occupied region and leave a blank cube in the rendered image, as shown in~\cref{fig:occ_update}.

\section{Experiments}
We demonstrate the efficient on-the-fly training capability of our method on both synthetic and real-world novel view synthesis datasets. We have explained in~\cref{problem_formulation} that we focus on scenes captured by multi-view forward-facing cameras since it is the most realistic setting in streaming applications. For the synthetic dataset, we evaluate on the widely used D-NeRF \cite{pumarola2020dnerf} dataset. However, the original D-NeRF dataset is captured with unrealistic teleporting cameras, and thus we render a forward-facing version using Blender for training and testing instead. This dataset will be released and more details are included in the supplementary. For the real-world dataset, we evaluate on the MeetRoom \cite{li2022streamrf} and DyNeRF \cite{li2022dynerf} datasets, which are both captured with multi-view forward-facing cameras. All results of our method are reported for models trained with a single RTX3090 GPU. Some rendered videos are included in the supplementary. 

\subsection{Synthetic Dataset}
On the synthetic dataset, we implement our method on top of the TiNeuVox~\cite{TiNeuVox} and evaluate the improvements in on-the-fly training speed and rendering quality. We also compare the performance of our method against the reported results of many other baseline models as shown in~\cref{tab:quantitative_dnerf}, but most of these models are trained jointly instead of trained on-the-fly. Our model is trained on the first frame for 200 iterations (for around 2 seconds), and then optimized for just 10 iterations per frame on-the-fly. Compared to the baseline methods, our model has significantly faster training  speed (12 FPS) and rendering speed(15 FPS), while achieving a superior rendering quality measured in PSNR and LPIPS~\cite{zhang2018lpips}. Our model can be trained and rendered at a total of 6.78 FPS, which makes it possible to be used for many interactive applications. 

\begin{table}[]
\begin{threeparttable}
\begin{tabular}{l|ccc|cc}
\toprule
& \multicolumn{3}{c|}{Frames Per Second (FPS)$\uparrow$} & \multicolumn{2}{c}{Render Quality} \\
\multicolumn{1}{c|}{Method}     & Train         & Render        & Total     
& PSNR$\uparrow$  & LPIPS$\downarrow$ \\ \hline
D-NeRF\cite{pumarola2020dnerf}*     & 0.0013    & NA         & \textless{}0.0013 & 30.50 & 0.07  \\
K-Plane\cite{kplanes_2023}*    & 0.04      & NA         & \textless{}0.04   & 31.67 & NA    \\
TiNeuVox-S\cite{TiNeuVox}* & 0.21      & NA         & \textless{}0.21   & 30.75 & 0.07  \\
TiNeuVox-B\cite{TiNeuVox}* & 0.08      & NA         & \textless{}0.08   & 32.67 & \textbf{0.04}  \\
TempInterp\cite{park2023temporal}* & 0.21      & NA         & \textless{}0.21   & 29.84 & 0.06  \\
TiNeuVox\cite{TiNeuVox}    & 1.27      & 2.80       & 0.87              & 30.57 & 0.07  \\
Ours        & \textbf{12.11}     & \textbf{15.68}      & \textbf{6.78}              & \textbf{32.87} & \textbf{0.04} \\
\bottomrule
\end{tabular}

\begin{tablenotes}
  \footnotesize
  \item *Result reported for joint training instead of on-the-fly training. NA if rendering speed not reported.
\end{tablenotes}
\caption{Quantitative results of speed and novel view synthesis qualities on the D-NeRF dataset.}
\label{tab:quantitative_dnerf}
\end{threeparttable}
\end{table}

\begin{figure}
    \includegraphics[width=\linewidth]{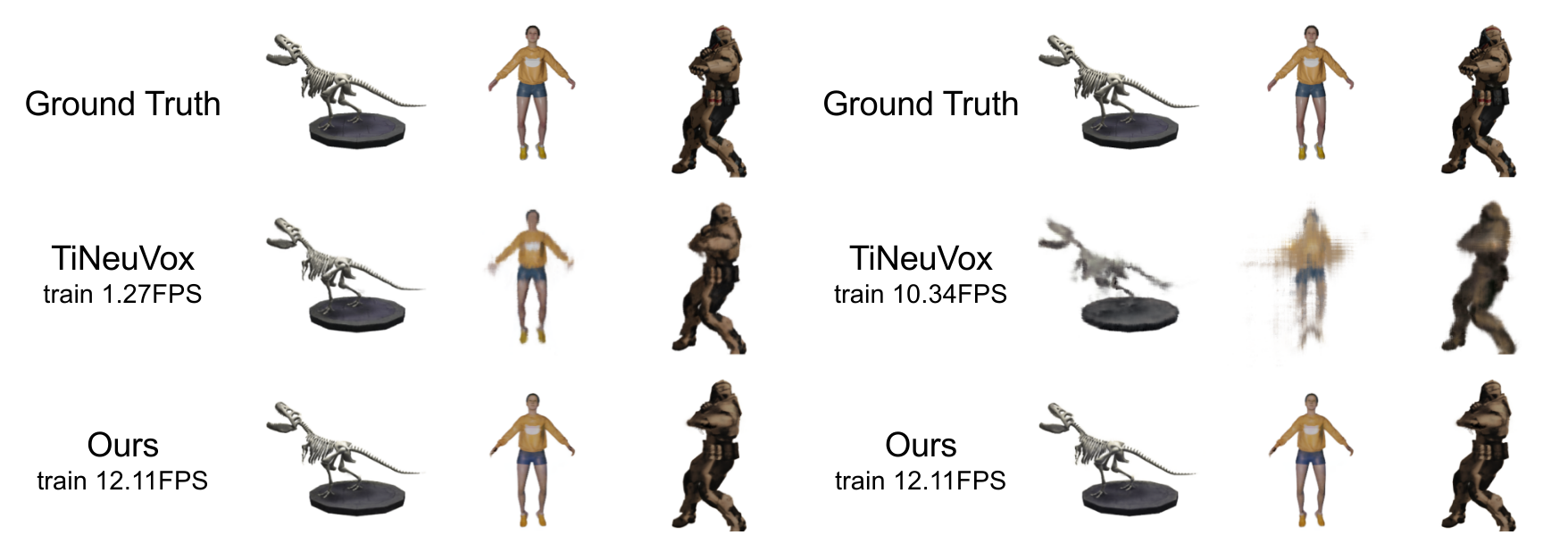}
    \caption{Qualitative results of our model compared to TiNeuVox on the D-NeRF dataset, by training to a comparable rendering quality(left) and training under a similar time constraint.}
    \vspace{-5mm}
    \label{fig:qualitative_synthetic}
\end{figure}

\begin{figure}[!ht]
\begin{floatrow}
\ffigbox[\FBwidth]
  {\includegraphics[width=6cm]{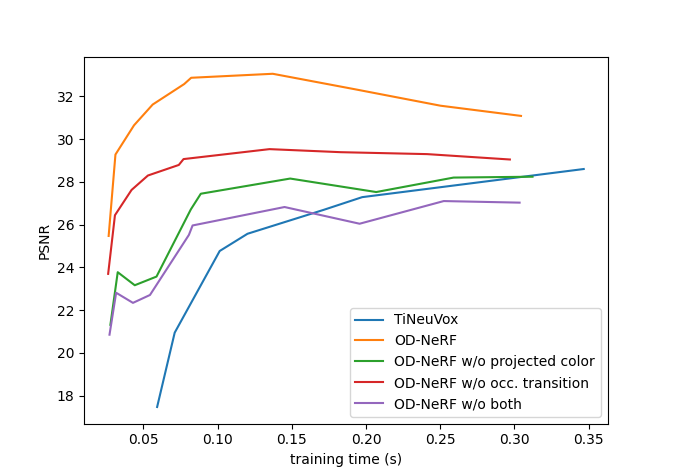}}
  {\caption{Ablation comparison of rendering quality with different training time.}\label{fig:ablation_curve}}
\floatbox{table}[\FBwidth]
  {
    \resizebox{0.45\textwidth}{!}{
        \begin{tabular}{l|ccc|cc}
        \toprule
        & \multicolumn{3}{c|}{Speed (FPS)$\uparrow$} & \multicolumn{2}{c}{Render Quality} \\
        \multicolumn{1}{c|}{Method}     & Train         & Render        & Total     
        & PSNR$\uparrow$  & LPIPS$\downarrow$ \\ \hline
        TiNeuVox\cite{TiNeuVox}           & 1.27      & 2.80       & 0.87      & 30.57 & 0.07  \\
        Ours               & 12.11     & 15.68      & 6.78      & \textbf{32.87} & \textbf{0.04}  \\
        No Proj. Color & 11.61     & 17.64      & 6.94      & 27.44 & 0.09  \\
        No Occ. Trans.  & \textbf{13.38}     & 17.75      & 7.57      & 29.06 & 0.05  \\
        No Both            & 12.40     & \textbf{20.70}      & \textbf{7.71}      & 25.96 & 0.10  \\
        \bottomrule
        \end{tabular}
    }
    \vspace{0.7cm}
    }
  {\caption{Quantitative ablation study of each component proposed for our model.}\label{tab:ablation}}
\end{floatrow}
\end{figure}

As shown in~\cref{fig:qualitative_synthetic}, we also demonstrate some qualitative results of our method compared to TiNeuVox\cite{TiNeuVox}. We compare the training speed and rendering quality when trained to a comparable quality or under similar time constraints. Our model can achieve comparable novel view synthesis quality with significantly faster on-the-fly training, or render with much superior quality under the same time constraint. We present an ablation study in~\cref{tab:ablation} to better analyze the effectiveness of the projected color guidance component and occupancy grid transition component proposed. We remove the projected color input and the occupancy transition function one by one for evaluation. All ablation models are trained for the same 10 iterations per frame. The ablation results suggest that both components contribute towards a fast convergence during on-the-fly training. The occupancy transition slightly reduces the average training speed as more points are sampled at the start of per-frame optimization due to the transited occupancy grid, but significantly improves the quality.

To better compare the performance of our proposed methods and the baseline model under different time constraints, we illustrate the plot showing their rendering quality (PSNR) against their training time in~\cref{fig:ablation_curve}. Our proposed OD-NeRFs perform significantly better than the TiNeuVox~\cite{TiNeuVox} baseline when at very strict time constraints. As the training time increases, the rendering quality gap between our method and the baseline reduces. It is worth noting that the performance of the proposed model deteriorates slightly when given more than around 0.15 seconds per frame to train, possibly caused by over-fitting to the last frame. 

\thisfloatsetup{floatwidth=0.8\paperwidth,rowfill=yes,margins=hangboth,heightadjust=all,valign=b}


\vspace{-3mm}
\subsection{Real-World Datasets}
For the real-world datasets, our model is trained on the first frame for 6000 iterations and later 100 iterations per frame. We compare the training speed and rendering quality of our method implemented on top of the NerfAcc \cite{li2023nerfacc} implementation of InstantNGP \cite{mueller2022instant} against various baseline models used for dynamic and static NeRFs. Although some of the models (\emph{e.g.} StreamRF \cite{li2022streamrf}) do not claim the on-the-fly training ability, they are compatible with the on-the-fly training proposed in our paper. As shown in~\cref{tab:quantitative_real}, our model can be trained on-the-fly significantly faster than the baseline models while maintaining a similar rendering quality. 

\begin{table}[]
\centering
\begin{adjustbox}{center}
\begin{tabular}{lccccllcccc}
\multicolumn{5}{c}{MeetRoom}                                                                                  &  & \multicolumn{5}{c}{DyNeRF}                                                                                         \\ \cline{1-5} \cline{7-11} 
                                & \multicolumn{3}{|c|}{Time(seconds per frame)$\downarrow$} & & & & \multicolumn{3}{|c|}{Time(seconds per frame)$\downarrow$} & \\
\multicolumn{1}{l|}{Method}     & Train & Render & \multicolumn{1}{c|}{Total} & PSNR$\uparrow$  &  & \multicolumn{1}{l|}{Method}     & Train & Render & \multicolumn{1}{c|}{Total}      & PSNR$\uparrow$  \\ \cline{1-5} \cline{7-11} 

\multicolumn{1}{l|}{-}          & -              & -               & \multicolumn{1}{c|}{-}             &   -   &  & \multicolumn{1}{l|}{DyNeRF\cite{li2022dynerf}*}      & 15600         & 67             & \multicolumn{1}{c|}{15667}              & 29.58 \\
\multicolumn{1}{l|}{JaxNeRF\cite{jaxnerf2020jaxnerf}*}   & 28380         & 40             & \multicolumn{1}{c|}{28420}         & 27.11 &  & \multicolumn{1}{l|}{K-Plane\cite{kplanes_2023}*}   & 21.6          & NA             & \multicolumn{1}{c|}{\textgreater{}21.6} & \textbf{31.63} \\
\multicolumn{1}{l|}{Plenoxels\cite{yu2021plenoxels}}  & 840           & 0.1            & \multicolumn{1}{c|}{840}           & 27.15 &  & \multicolumn{1}{l|}{Plenoxels\cite{yu2021plenoxels}}  & 1380          & 0.12           & \multicolumn{1}{c|}{1380}               & 28.68 \\
\multicolumn{1}{l|}{LLFF\cite{suhail2022light}}       & 180           & \textbf{0.0003}         & \multicolumn{1}{c|}{180}           & 22.88 &  & \multicolumn{1}{l|}{LLFF\cite{suhail2022light}}       & 480           & \textbf{0.004}          & \multicolumn{1}{c|}{480}                & 23.23 \\
\multicolumn{1}{l|}{StreamRF\cite{li2022streamrf}}   & 10.2          & 0.1            & \multicolumn{1}{c|}{10.3}          & 26.72 &  & \multicolumn{1}{l|}{StreamRF\cite{li2022streamrf}}   & 15.0          & 0.1            & \multicolumn{1}{c|}{15.1}               & 28.26 \\
\multicolumn{1}{l|}{InstNGP\cite{mueller2022instant}} & 12.9          & 0.8            & \multicolumn{1}{c|}{13.6}          & 22.82 &  & \multicolumn{1}{l|}{InstNGP\cite{mueller2022instant}} & 12.6          & 1.0            & \multicolumn{1}{c|}{13.6}               & 24.30 \\
\multicolumn{1}{l|}{Ours}       & \textbf{2.3}           & 1.2            & \multicolumn{1}{c|}{\textbf{3.5}}           & \textbf{27.32} &  & \multicolumn{1}{l|}{Ours}       & \textbf{2.7}           & 2.4            & \multicolumn{1}{c|}{\textbf{5.0}}                & 27.52 \\ \cline{1-5} \cline{7-11} 
\end{tabular}
\end{adjustbox}
\footnotesize *Result reported for joint training instead of on-the-fly training. NA if rendering speed not reported.
\caption{Quantitative results on real-world MeetRoom and DyNeRF dataset. }
\label{tab:quantitative_real}
\end{table}

\begin{figure}
    \centering
    \includegraphics[width=0.9\linewidth]{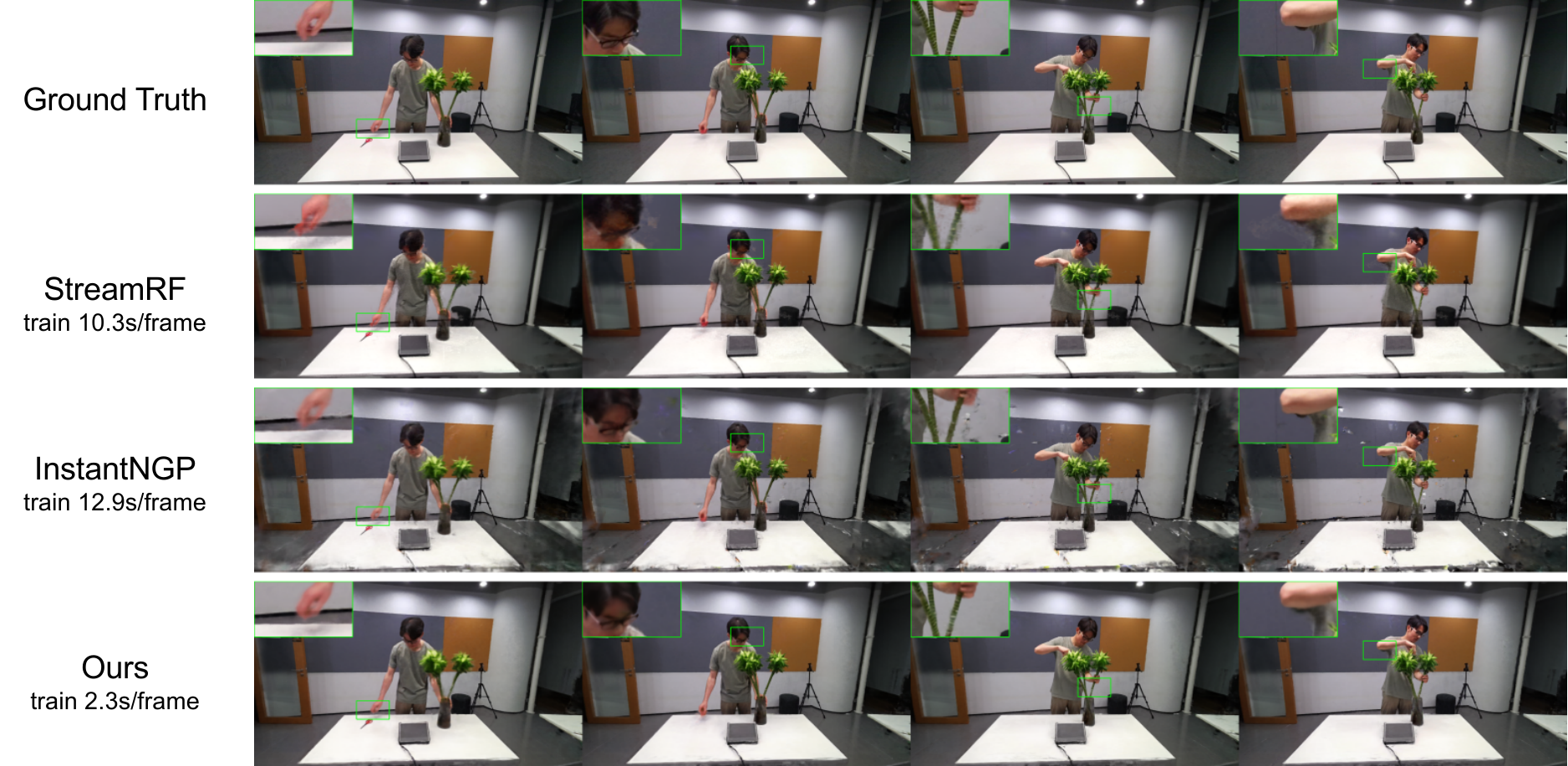}
    \caption{Qualitative results of models with different training speed on the MeetRoom dataset.}
    \vspace{-2mm}
    \label{fig:qualitative_trimming}
\end{figure}

\begin{figure}
    \centering
    \includegraphics[width=0.9\linewidth]{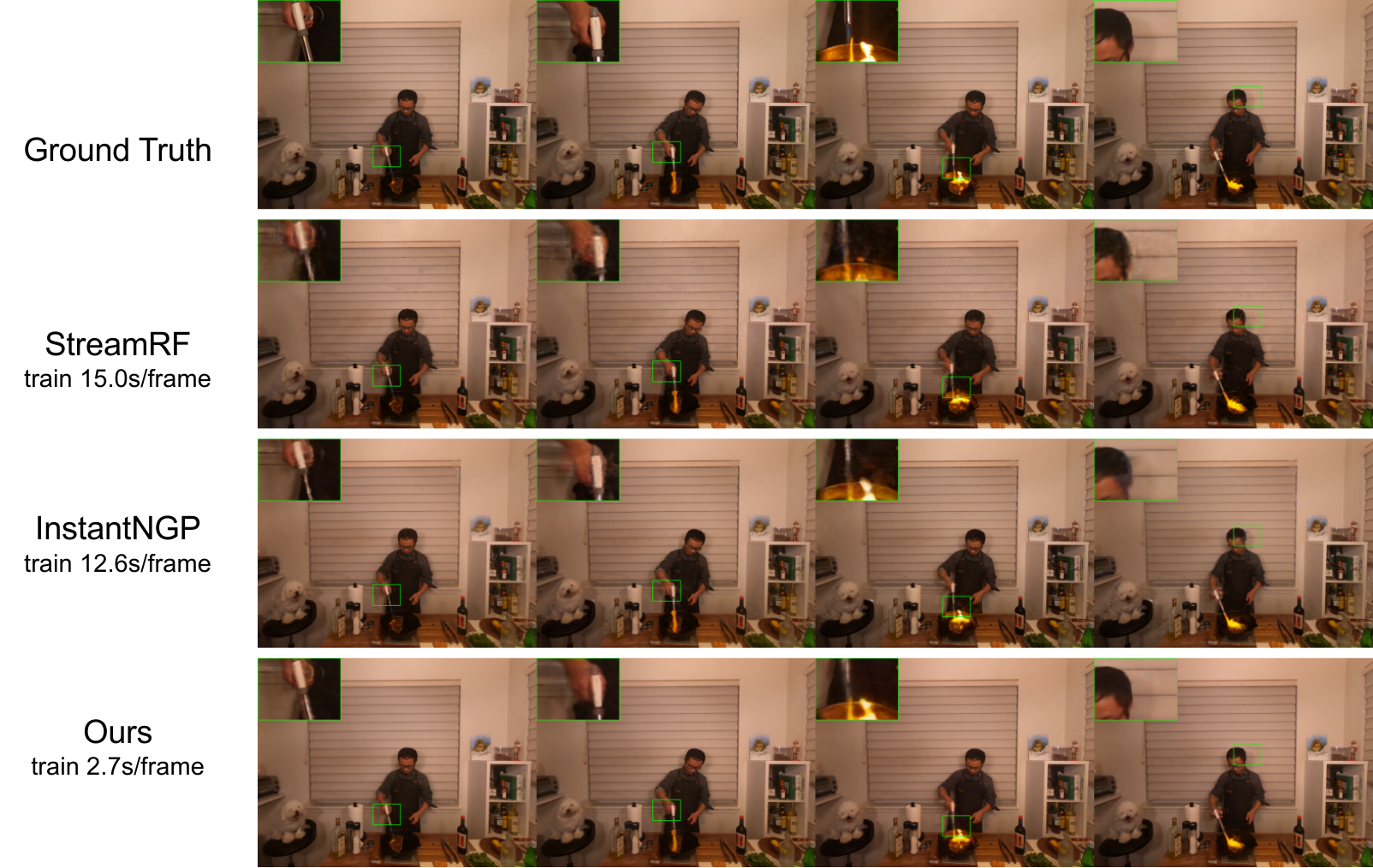}
    \caption{Qualitative results of models with different training speed on the DyNeRF dataset.}
    \vspace{-2mm}
    \label{fig:qualitative_steak}
\end{figure}

We also present some of the qualitative results of our model compared to StreamRF \cite{li2022streamrf} and InstantNGP \cite{mueller2022instant} as shown in~\cref{fig:qualitative_trimming} on MeetRoom \cite{li2022streamrf} dataset and in~\cref{fig:qualitative_steak} on DyNeRF \cite{li2022dynerf} dataset. Our model is free of many common artifacts present in the rendered results of the baseline model while maintaining a significantly faster on-the-fly training speed. 

\section{Limitations}
\vspace{-2mm}
The implicit correspondence of our projected color-guided NeRF relies on the relative invariance of the projected color of a point. However, this invariance can be violated with specular surfaces and occluded points. It may be possible to filter out the outlier projected colors caused by specularity and occlusion, or explicitly detect occlusion. However, this process may incur significant computation costs and can be further analyzed in future works. 

\section{Conclusion}
\vspace{-2mm}
We introduced a new on-the-fly dynamic NeRF training setting, where the radiance field is trained and rendered frame by frame. To tackle the efficient on-the-fly training challenge, we propose a projected color-guided dynamic NeRF conditioned on the spatial and color input to efficiently optimize the radiance field with implicit point correspondence. We also propose a transition and update function to the occupancy grid for efficient point sampling in space. The experiment results in both synthetic and real-world datasets indicate the superior on-the-fly training speed of our method while maintaining a comparable rendering quality. 

{\small
\bibliographystyle{plain}
\bibliography{egbib}
}


%

\clearpage
\appendix
\section*{Appendices}
\addcontentsline{toc}{section}{Appendices}

\section{Qualitative Result Videos}
We include a few videos rendered by our model and baselines in the supplementary zip file. 

\section{Implementation Details}
We implement our model on top of the TiNeuVox\cite{TiNeuVox} for the synthetic dataset and the InstantNGP\cite{mueller2022instant} for the real-world dataset. We describe the changes we have made to the models in this section.
\subsection{Synthetic Dataset Model}
As we have mentioned in the main paper, we remove the temporal components of the model because of its poor extrapolation capability. More specifically, the temporal deformation model and the temporal information enhancement are removed. Instead, the mean and variance projected color of the sampled point is concatenated with its spatial feature as the input to the NeRF Multi-Layer Perceptron(MLP). We also replace the multi-scale voxel used in TiNeuVox\cite{TiNeuVox} with the hash voxel used in InstantNGP\cite{mueller2022instant}, implemented with tiny-cuda-nn\cite{tiny-cuda-nn}. We observe that this hashed voxel can better capture the details, but converge slower than the original multi-scale voxel. Hence we added the 2-second warm-up for the first frame as mentioned in the main paper. The rest of the model structure is following the TiNeuVox-S version published. 

Since the original TiNeuVox sample uniformly along the ray instead of sampling based on the occupancy grid, we implement a rejection sampling based on our transited and updated occupancy grid. The rejection sampling filters the uniform samples based on the occupancy grid and a fixed interval. The $i$th sample $\mathbf{x}_i$ along the ray is rejected if the occupancy value from the occupancy grid $\sigma_{occ}(\mathbf{x}_i)$ is smaller than a density threshold $\sigma_{min}$ and it is not on a fixed interval $R$:
\begin{equation}
    reject\; \mathbf{x}_i\; if\quad (\sigma_{occ}(\mathbf{x}_i) < \sigma_{min})\; and\; \neg(i\;mod\; R \equiv R//2).
\end{equation}
We fix the interval $R$ with a value of $20$, and gradually decrease the threshold $\sigma_{min}$ over the optimization process from $1$(at the frame $1$) to $0.05$(at frame $10$) for better convergence at the start. 

\subsection{Real-world Dataset Model}
The NerfAcc\cite{li2023nerfacc} implementation of InstantNGP\cite{mueller2022instant} is used as the code base for our implementation. We concatenate the mean and variance of the projected colors of the sampled point with its spatial feature queried from the hash voxel grid, before inputting them into the NeRF MLP. We also notice that the NerfAcc\cite{li2023nerfacc} implementation of InstantNGP\cite{mueller2022instant} does not converge very well on the forward facing dynamic dataset of MeetRoom\cite{li2022streamrf} and DyNeRF\cite{li2022dynerf}, even for the first frame static scene. It could be because of the large planer background with constant colors, like walls and tables. Hence, we implement a simple depth smoothness regularization based on patch sampling. For any $3 \times 3$ patch of ray sampled, the depth regularization loss is calculated as:
\begin{equation}
    \mathcal{L}_{depth} = \frac{\mathrm{std}(d_{far} / d_{3x3})}{\mathrm{std}(c_{3x3})},
\end{equation}
where $\mathrm{std}$ represents the standard deviation, $d_{3x3}$ represents the depth values of the patch, $c_{3x3}$ represents the ground truth color of the patch and $d_{far}$ represents the far plane depth. This depth smoothness loss penalizes local large inverse depth variation when the color variation is small. The loss is added to the total loss with a weight of $1e^{-4}$ for MeetRoom\cite{li2022streamrf} dataset and $1e^{-6}$ for DyNeRF\cite{li2022dynerf} dataset. 

Since the InstantNGP\cite{mueller2022instant} model already has a sampling strategy based on the occupancy grid, we only update the occupancy grid itself during the training and do not change the sampling strategy itself. 

\section{Modifications to D-NeRF Dataset}
As we have mentioned in main paper, we use a multi-view forward facing camera version of the D-NeRF\cite{pumarola2020dnerf} dataset. Since the original D-NeRF dataset does not release the blender file, we use the publicly available models and animations to render with multi-forward facing cameras. However, we could not find the scene ``BouncingBalls'' used in the dataset, hence train and render with a TiNeuVox-B\cite{TiNeuVox} instead. 

Similar to the real-world forward facing dataset, this multi-view forward facing version of D-NeRF dataset has 12 static training cameras and 1 static test cameras. The test camera is in the center and the training camera arranged in 3 rows and 4 columns with a $40^\circ$ spread. The cameras all look at the center of the synthetic model. 

\section{Additional Ablation Results}
To better analyze the effectiveness of the projected color mean and variance individually, we illustrate an additional ablation study using only the mean or variance of the projected color in~\cref{tab:ablation_mean_var}. This quantitative comparison suggests that the mean of projected color increases the performance significantly. Using the projected color variance further improves the performance slightly. 

\begin{table}[]
    \centering
        \begin{tabular}{l|ccc|cc}
        \toprule
        & \multicolumn{3}{c|}{Speed (FPS)$\uparrow$} & \multicolumn{2}{c}{Render Quality} \\
        \multicolumn{1}{c|}{Method}     & Train         & Render        & Total     
        & PSNR$\uparrow$  & LPIPS$\downarrow$ \\ \hline
        Full OD-NeRF               & \textbf{12.11}     & 15.68      & \textbf{6.78}      & \textbf{32.87} & \textbf{0.04}  \\
        Var Only  & 9.70      & 12.90      & 5.33      & 28.15 & 0.08  \\
        Mean Only & 11.59     & \textbf{16.11}      & 6.65      & 32.22 & 0.04 \\
        No Mean/Var & 10.59     & 15.61      & 6.17      & 25.53 & 0.09 \\
        \bottomrule
        \end{tabular}
    \caption{Quantitative ablation study of each component proposed for our model.}\label{tab:ablation_mean_var}
\end{table}

\section{Additional Qualitative Results}
\begin{figure}
    \centering
    \includegraphics[width=\textwidth]{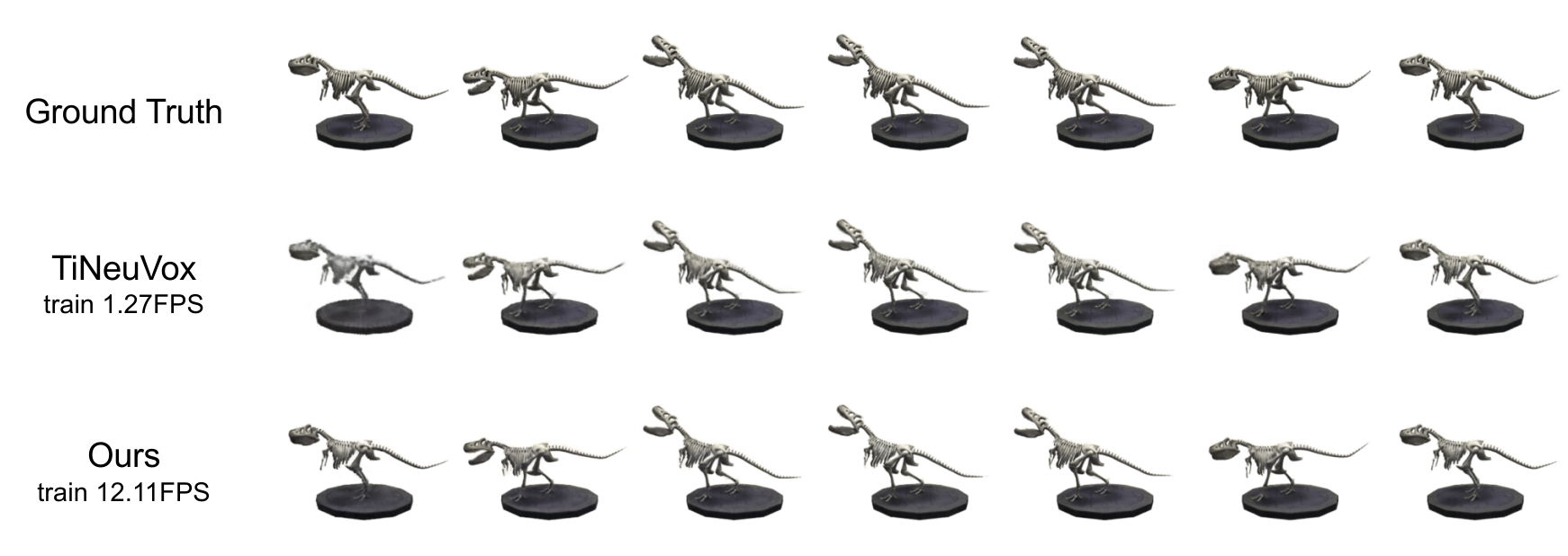}
    \caption{Qualitative results of the ``T-Rex'' scene in the D-NeRF dataset, where the two models are trained to a \textbf{comparable rendering quality}. }
    \label{fig:qualitative_trex_same_quality}
\end{figure}

\begin{figure}
    \centering
    \includegraphics[width=\textwidth]{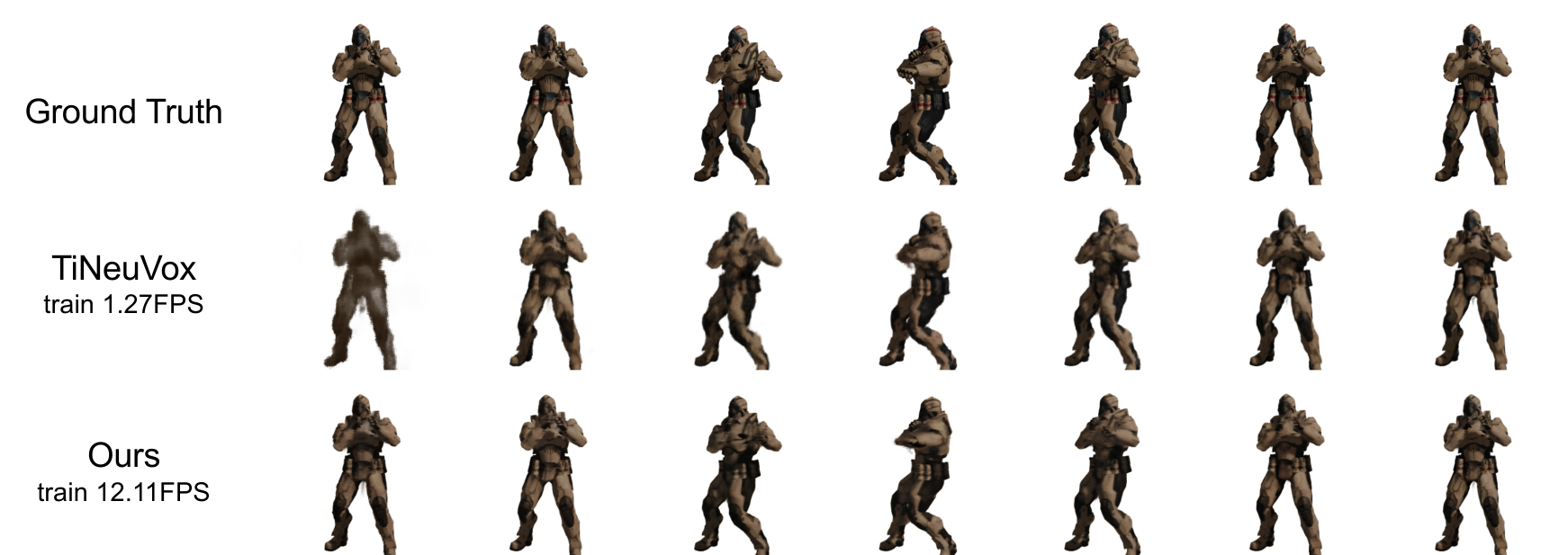}
    \caption{Qualitative results of the ``Hook'' scene in the D-NeRF dataset, where the two models are trained to a \textbf{comparable rendering quality}. }
    \label{fig:qualitative_hook_same_quality}
\end{figure}

\begin{figure}
    \centering
    \includegraphics[width=\textwidth]{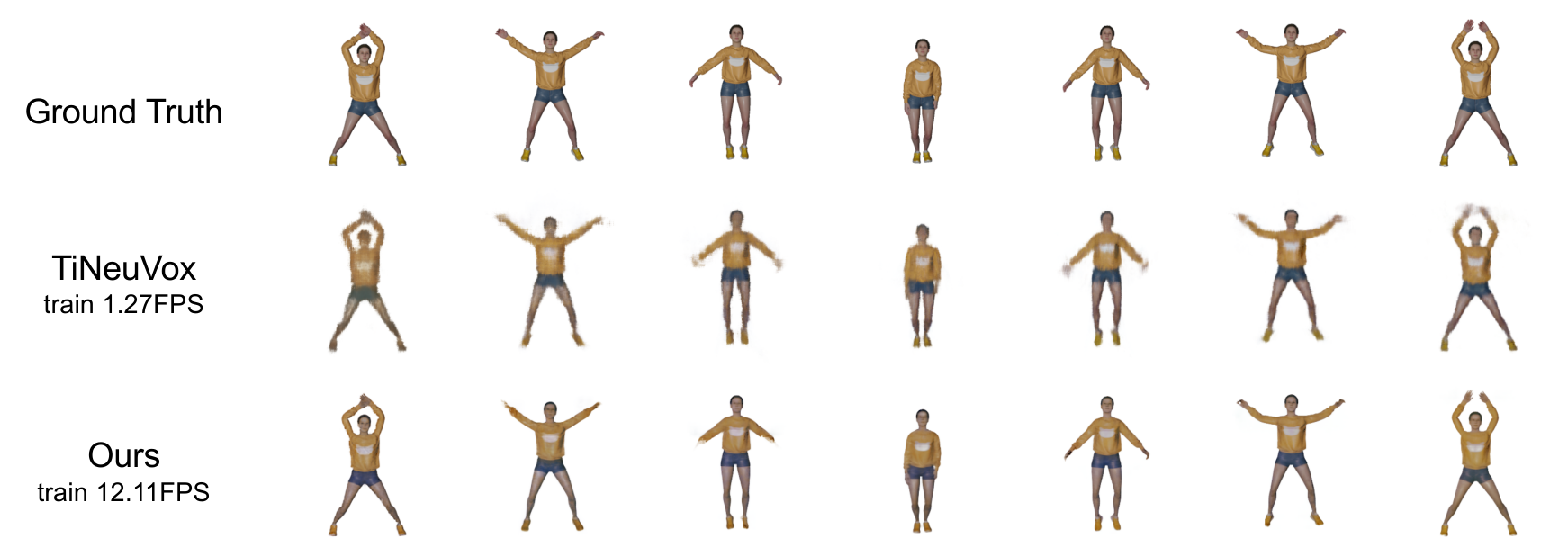}
    \caption{Qualitative results of the ``Jumping-Jack'' scene in the D-NeRF dataset, where the two models are trained to a \textbf{comparable rendering quality}. }
    \label{fig:qualitative_jumpingjack_same_quality}
\end{figure}

\begin{figure}
    \centering
    \includegraphics[width=\textwidth]{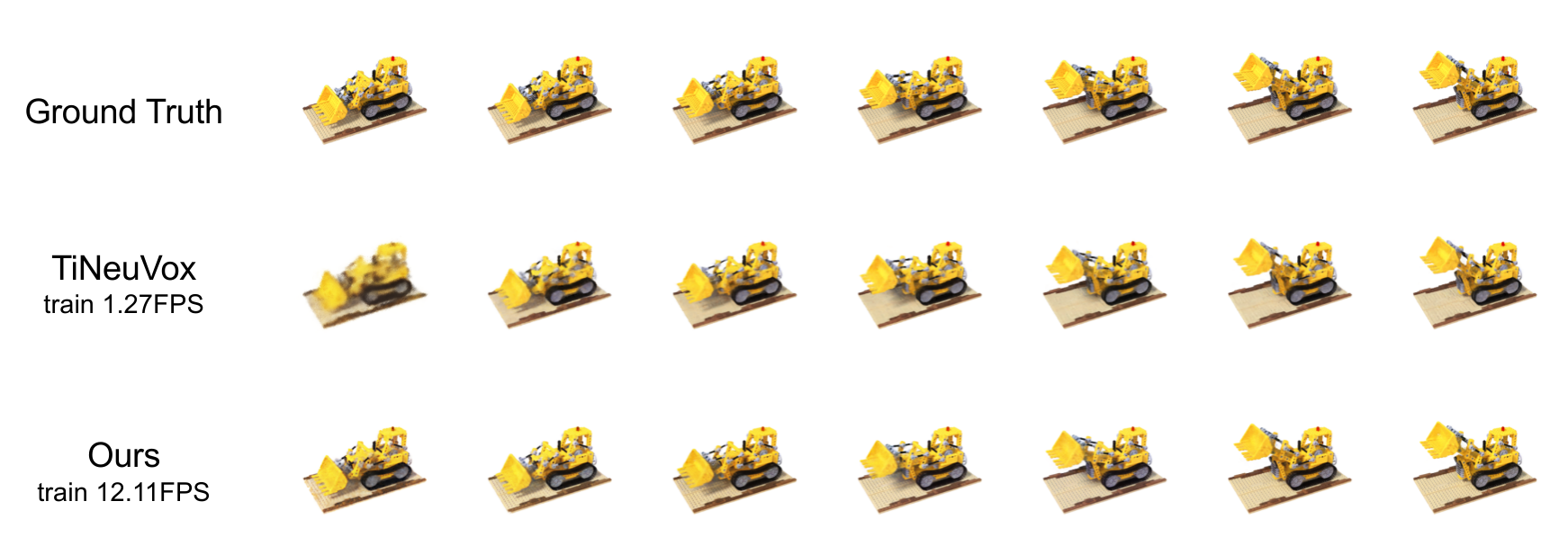}
    \caption{Qualitative results of the ``Lego'' scene in the D-NeRF dataset, where the two models are trained to a \textbf{comparable rendering quality}. }
    \label{fig:qualitative_lego_same_quality}
\end{figure}

\begin{figure}
    \centering
    \includegraphics[width=\textwidth]{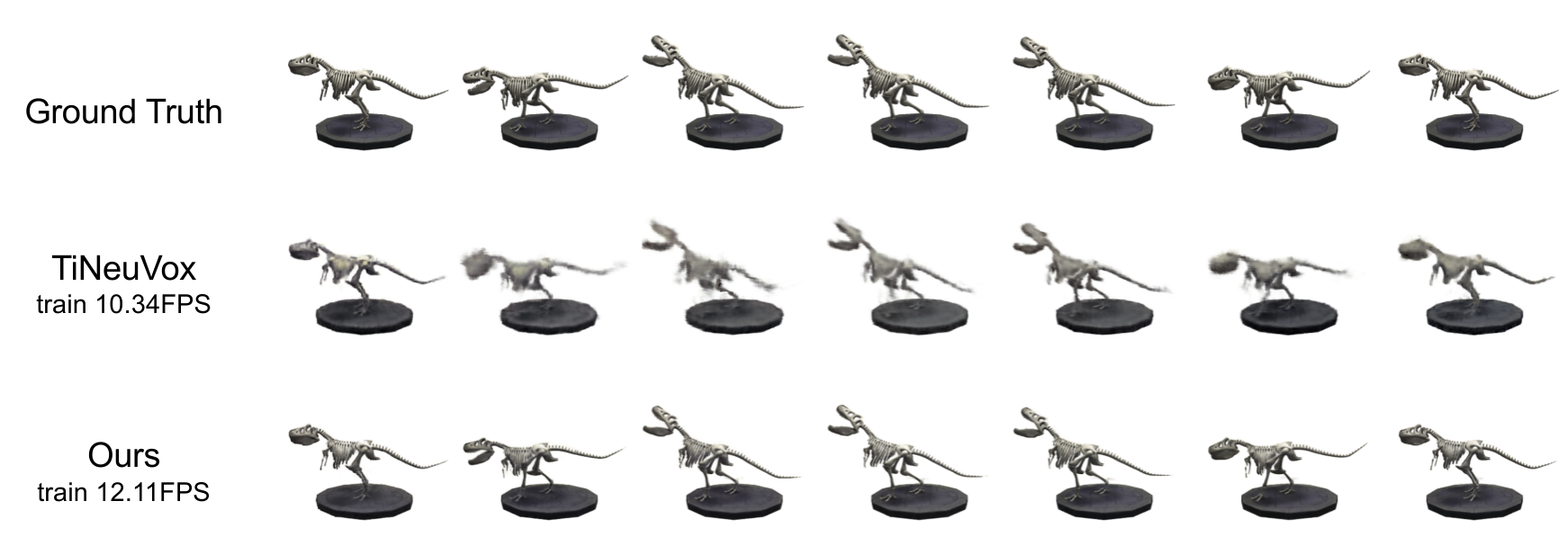}
    \caption{Qualitative results of the ``T-Rex'' scene in the D-NeRF dataset, where the two models are trained with a  \textbf{similar time constraint}. }
    \label{fig:qualitative_trex_same_time}
\end{figure}

\begin{figure}
    \centering
    \includegraphics[width=\textwidth]{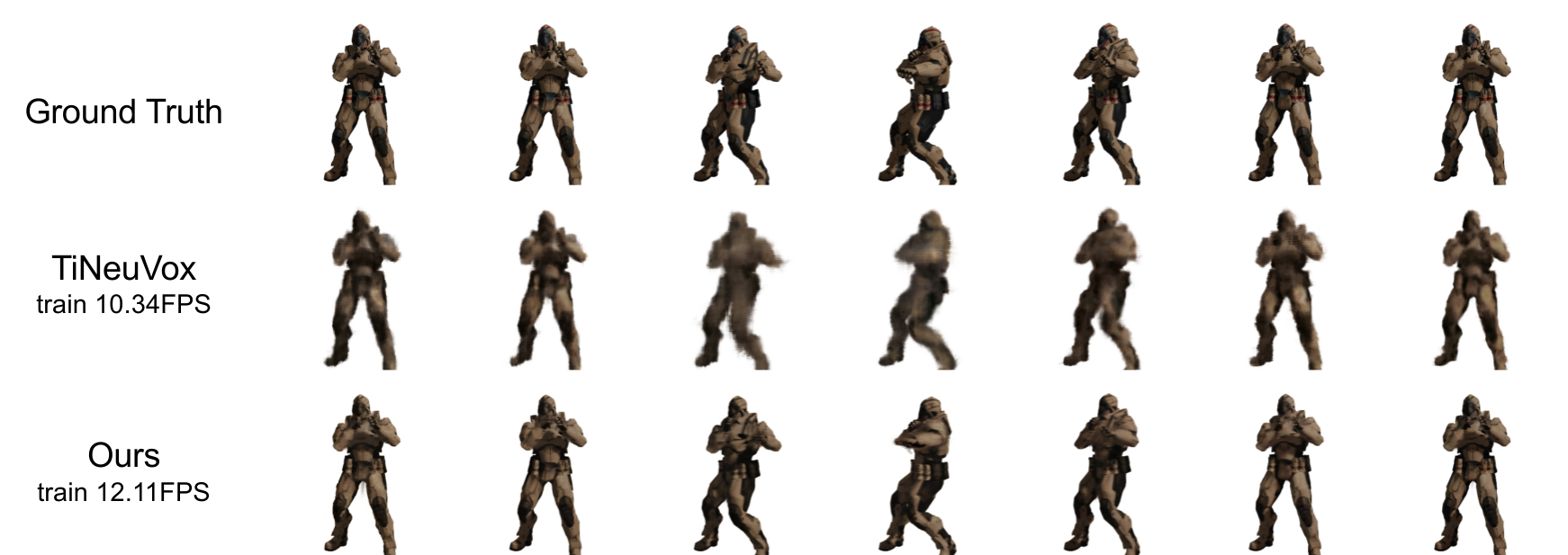}
    \caption{Qualitative results of the ``Hook'' scene in the D-NeRF dataset, where the two models are trained with a  \textbf{similar time constraint}. }
    \label{fig:qualitative_hook_same_time}
\end{figure}

\begin{figure}
    \centering
    \includegraphics[width=\textwidth]{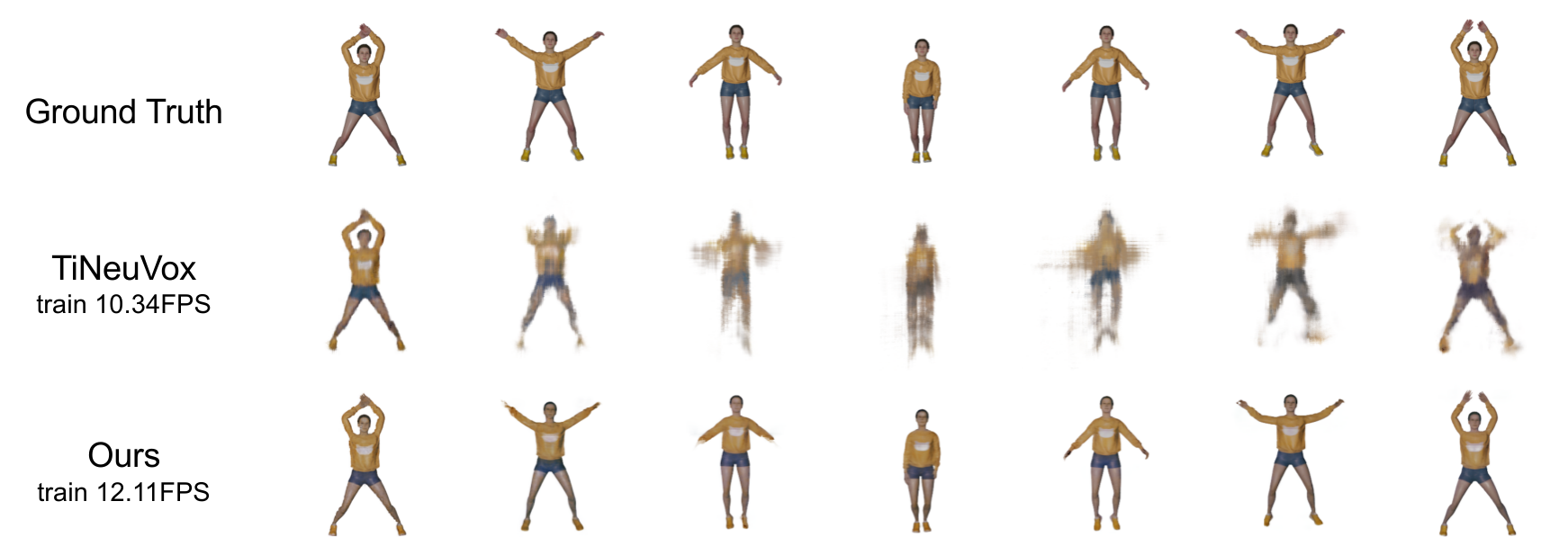}
    \caption{Qualitative results of the ``Jumping-Jack'' scene in the D-NeRF dataset, where the two models are trained with a  \textbf{similar time constraint}. }
    \label{fig:qualitative_jumpingjack_same_time}
\end{figure}

\begin{figure}
    \centering
    \includegraphics[width=\textwidth]{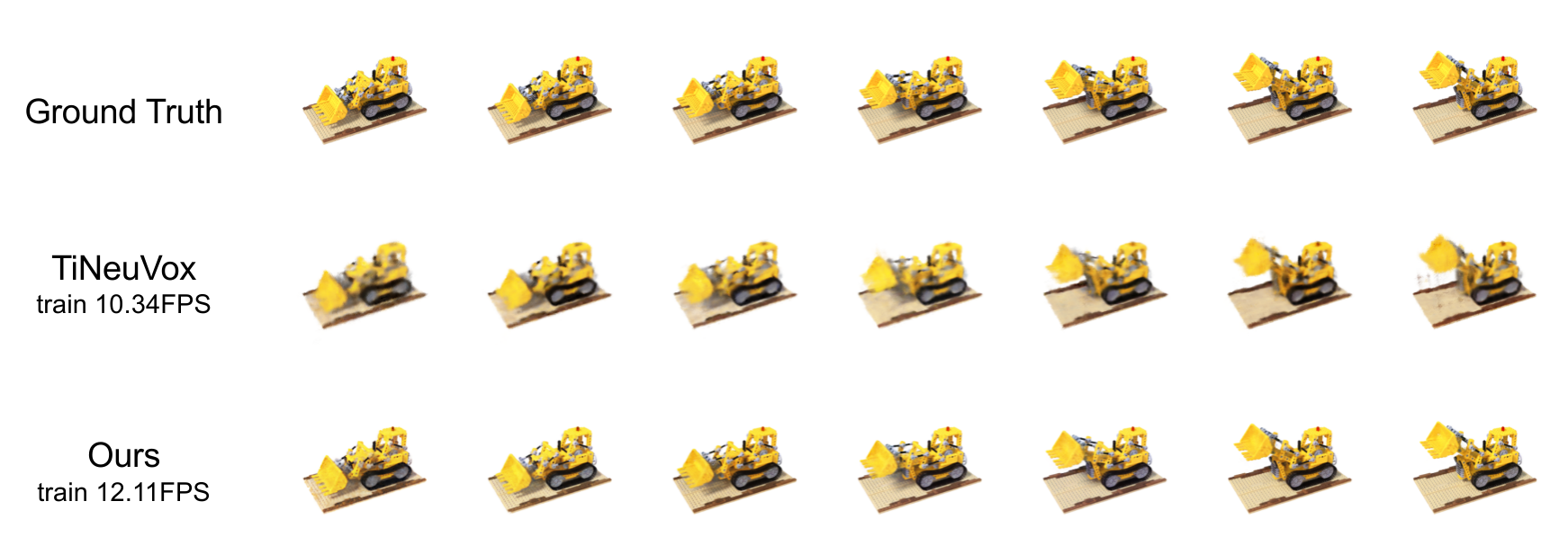}
    \caption{Qualitative results of the ``Lego'' scene in the D-NeRF dataset, where the two models are trained with a  \textbf{similar time constraint}. }
    \label{fig:qualitative_lego_same_time}
\end{figure}

\begin{figure}
    \centering
    \includegraphics[width=\textwidth]{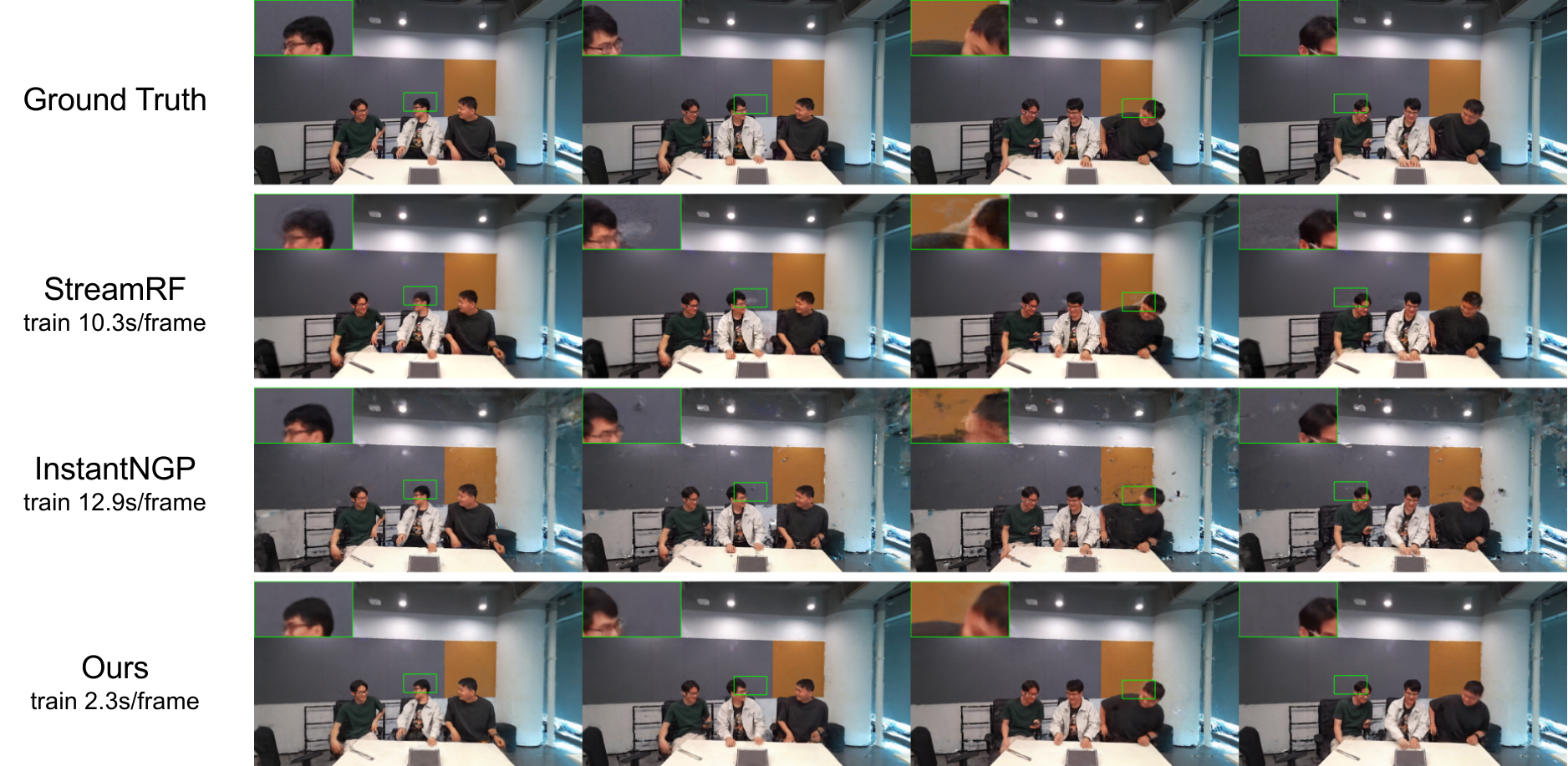}
    \caption{Qualitative results of the ``discussion'' scene in the MeetRoom dataset.}
    \label{fig:qualitative_discussion}
\end{figure}

\begin{figure}
    \centering
    \includegraphics[width=\textwidth]{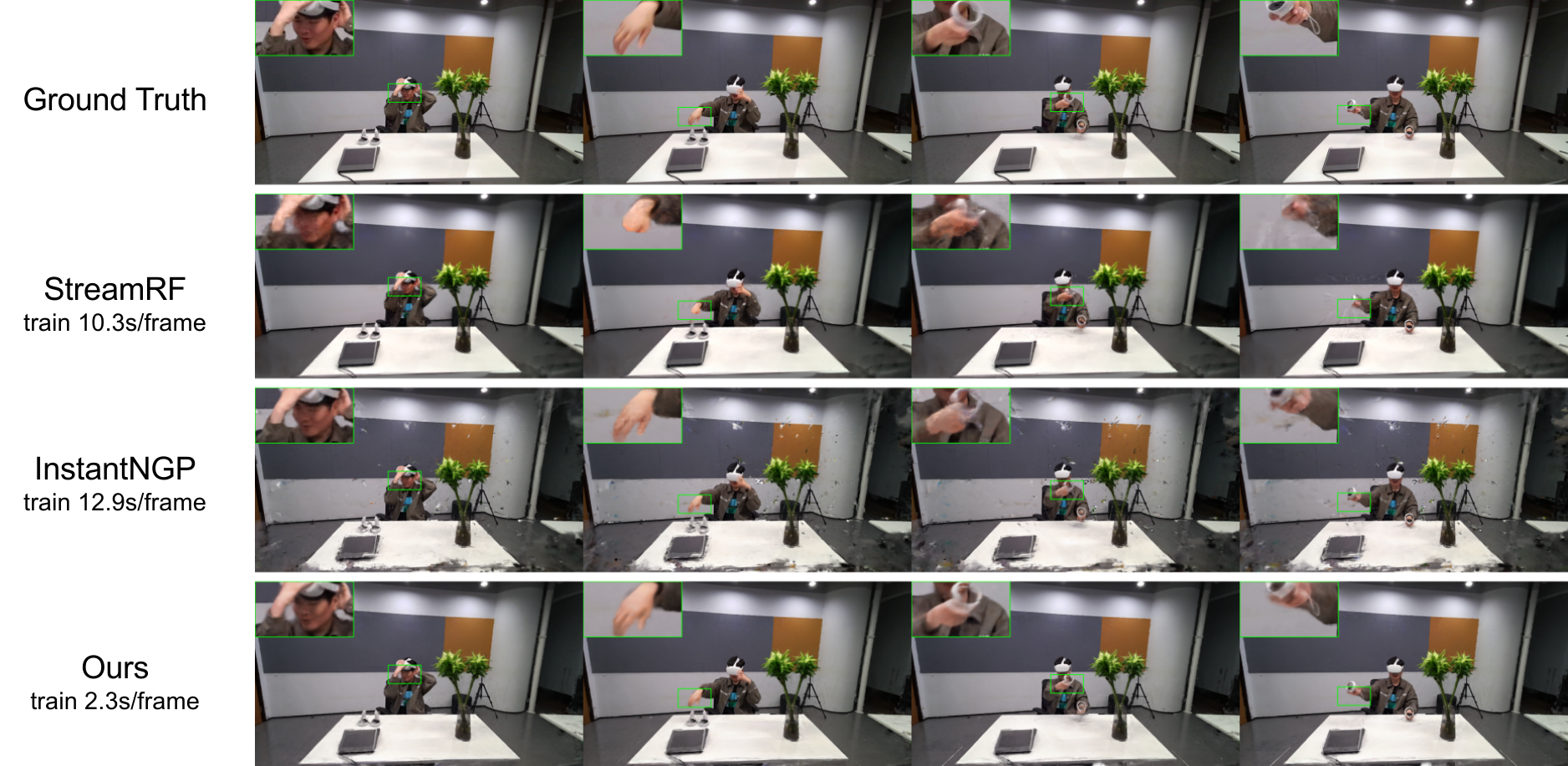}
    \caption{Qualitative results of the ``vrheadset'' scene in the MeetRoom dataset.}
    \label{fig:qualitative_vrheadset}
\end{figure}

We present some additional qualitative results for both the synthetic and real-world dataset. For the synthetic dataset, we present qualitative comparisons of models trained to a comparable rendering quality but different time(\cref{fig:qualitative_trex_same_quality}, \ref{fig:qualitative_hook_same_quality}, \ref{fig:qualitative_jumpingjack_same_quality} and \ref{fig:qualitative_lego_same_quality}), and trained under similar time constraint but with different rendering quality(\cref{fig:qualitative_trex_same_time}, \ref{fig:qualitative_hook_same_time}, \ref{fig:qualitative_jumpingjack_same_time} and \ref{fig:qualitative_lego_same_time}). For the real-world dataset, we show more rendering results of different scenes(\cref{fig:qualitative_discussion} and \ref{fig:qualitative_vrheadset}). 

\end{document}